\documentclass{Interspeech}

\interspeechcameraready

\title{GIA-MIC: Multimodal Emotion Recognition with Gated Interactive Attention and Modality-Invariant Learning Constraints}

\author[affiliation={1}]{Jiajun}{He}
\author[affiliation={1}]{Jinyi}{Mi}
\author[affiliation={2}]{Tomoki}{Toda}

\affiliation{Graduate School of Informatics}{Nagoya University}{Japan}
\affiliation{Information Technology Center}{Nagoya University}{Japan}
\email{\href{mailto:jiajun.he@g.sp.m.is.nagoya-u.ac.jp}{jiajun.he@g.sp.m.is.nagoya-u.ac.jp}, \href{mailto:mi.jinyi@g.sp.m.is.nagoya-u.ac.jp}{mi.jinyi@g.sp.m.is.nagoya-u.ac.jp},  tomoki@icts.nagoya-u.ac.jp}

\usepackage{comment}
\usepackage{amsmath}
\usepackage{graphicx}
\usepackage{multirow}
\usepackage{threeparttable}
\usepackage{graphicx}
\usepackage{booktabs}
\usepackage{multirow}
\usepackage[table]{xcolor}
\usepackage{enumitem}

\begin{document}

\maketitle

\begin{abstract}
\vspace{-2mm}

Multimodal emotion recognition (MER) extracts emotions from multimodal data, including visual, speech, and text inputs, playing a key role in human-computer interaction. Attention-based fusion methods dominate MER research, achieving strong classification performance. However, two key challenges remain: effectively extracting modality-specific features and capturing cross-modal similarities despite distribution differences caused by modality heterogeneity. To address these, we propose a gated interactive attention mechanism to adaptively extract modality-specific features while enhancing emotional information through pairwise interactions. Additionally, we introduce a modality-invariant generator to learn modality-invariant representations and constrain domain shifts by aligning cross-modal similarities. Experiments on IEMOCAP demonstrate that our method outperforms state-of-the-art MER approaches, achieving WA 80.7\% and UA 81.3\%.

\end{abstract}

\vspace{-1mm}
\section{Introduction}
\vspace{-1mm}
Multimodal emotion recognition (MER) aims to leverage information from multiple perceptual modalities, such as visual, acoustic, and textual expressions, to accurately identify human emotions. MER has broad applications in fields such as human-computer interaction and intelligent customer service \cite{tian2023semi}.

Although MER has demonstrated significant promise, it still faces two key challenges: effective feature extraction from each modality and robust multimodal fusion.
The first challenge lies in extracting meaningful modality-specific features. Early MER approaches primarily relied on training models from scratch, utilizing architectures such as recurrent neural networks (RNNs) \cite{liu2020group,LI2021114683}, convolutional neural networks (CNNs) \cite{guo2021representation,mi2024two}, and Transformers \cite{wu2024transformer,Waligora_2024_CVPR}. These methods improved emotion recognition by capturing latent patterns from video frames, audio spectrograms, and text sequences.
More recently, inspired by the success of pretrained models in other speech-related tasks, researchers have discovered that deep pretrained features offer more robust and generalizable representations than handcrafted features \cite{yang2024multi}. Consequently, various pretrained models have been adopted for MER, including visual-based models (e.g., CLIP \cite{radford2021clip, he24_interspeech}, DINO \cite{darcet2023vitneedreg}), acoustic-based models (e.g., HuBERT \cite{hsu2021hubert}, WavLM \cite{chen2022wavlm}), and text-based models (e.g., DeBERTa \cite{he2020deberta}, RoBERTa \cite{liu1907roberta}). These models have significantly enhanced MER performance by leveraging rich feature representations learned from large-scale datasets.

The second challenge is effectively integrating information across modalities. Attention mechanisms have played a pivotal role in MER research owing to their ability to selectively focus on emotion-relevant features. Traditional approaches often employ self-attention, where both the query and key originate from the same modality \cite{liu2022multi}. While effective, this method lacks direct cross-modal interaction, limiting its ability to fully exploit complementary information across modalities. To address this limitation, cross-attention mechanisms have been introduced, enabling one modality to attend to another by deriving queries and keys from different modalities, and utilizing values to retrieve the corresponding weighted information from the attended modality \cite{Praveen2024RecursiveJC}. Although cross-attention improves information exchange, it still struggles with modality heterogeneity, which can lead to inconsistencies and imbalances in feature alignment.

To address these challenges, we propose a novel method, GIA-MIC, which integrates gated interactive attention (GIA) for modality-specific representation (MSR) learning and modality-invariant learning constraints (MIC) to enhance modality-invariant representation (MIR) learning. Specifically, GIA facilitates adaptive extraction of modality-specific representations by dynamically regulating intermodal interactions, thereby capturing fine-grained emotional cues. Meanwhile, MIC encourages cross-modal alignment by constraining domain shifts between different modalities and learning modality-invariant representations through similarity-based constraints. We conducted extensive experiments on the Interactive Emotional Dyadic Motion Capture (IEMOCAP) dataset, and the results demonstrate the effectiveness of GIA-MIC, achieving new state-of-the-art (SOTA) performance in multimodal emotion recognition.
Our main contributions are as follows:
\vspace{-1mm}

\begin{itemize}
    \item We propose a GIA mechanism that dynamically regulates the importance of intermodal interactions, enabling the adaptive extraction of modality-specific representations.
     \item We introduce a modality-invariant generator (MIG) module along with MIC to learn robust modality-invariant representations. By capturing cross-modal similarities and enforcing constraints that reduce modality discrepancies, MIC enhances the generalization ability of the model, making it more resilient to domain shifts between different modalities and further improving MER performance.
    \item Our proposed GIA-MIC method outperforms state-of-the-art MER methods on the IEMOCAP dataset.  
\end{itemize}

\vspace{-3mm}
\section{Proposed Method}
\vspace{-2mm}
\begin{figure*}[ht]
    \centering
    \includegraphics[width=\linewidth]{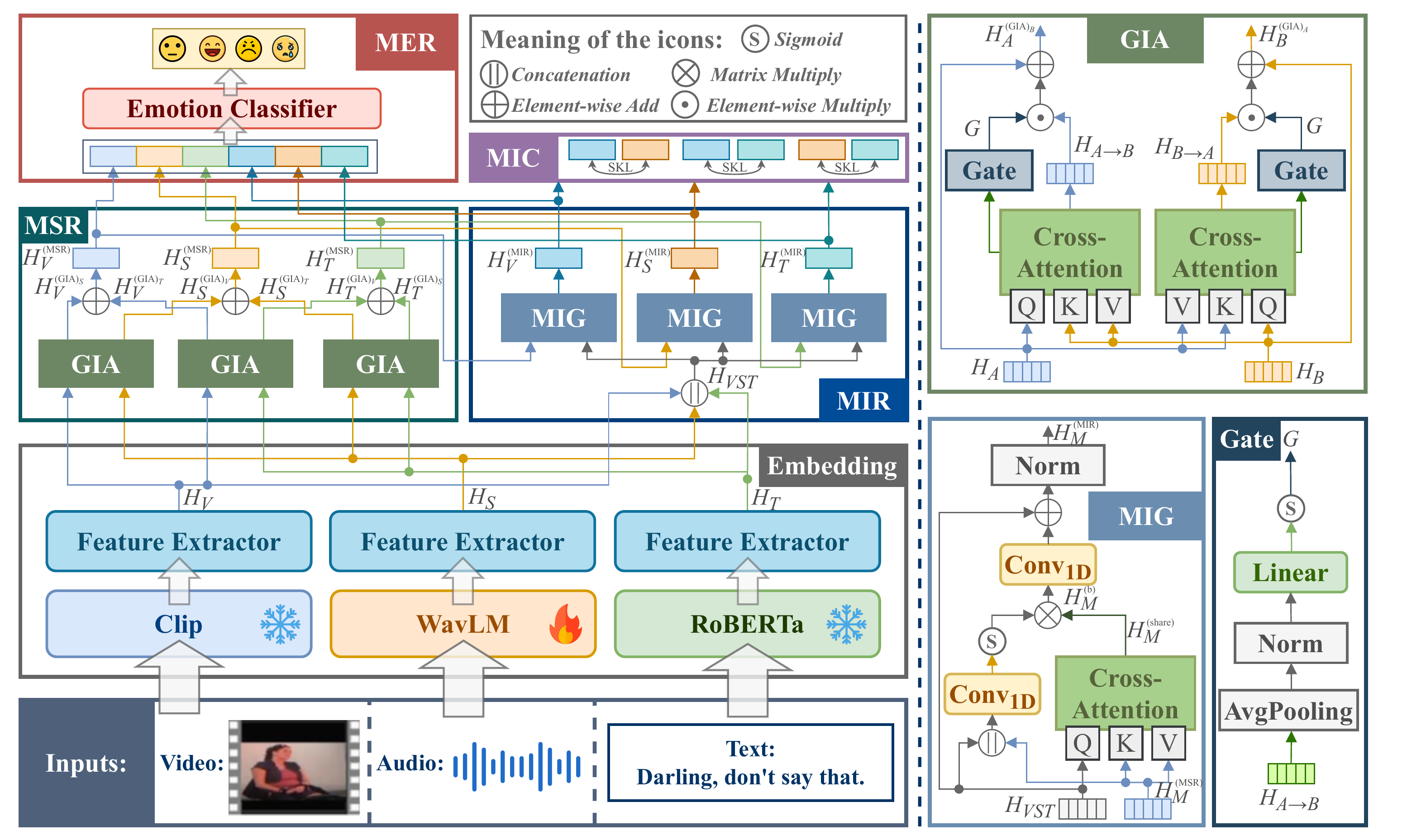}
    \vspace{-6mm}
    \caption{The overall architecture of our proposed method.}
    \label{fig:model}
    \vspace{-6mm}
\end{figure*}

\subsection{Problem Formulation}
\vspace{-2mm}
The MER task is defined as \( f(V, S, T) = L \), where \( V \), \( S \), and \( T \) represent the video, speech, and text modalities, respectively. The goal is to fuse these modalities for emotion classification, yielding \( L \in \{l_1, l_2, \dots, l_e\} \), where \( e \) is the number of emotion categories.
\vspace{-2mm}
\subsection{Embedding Module}
\vspace{-2mm}

We utilize pretrained encoders CLIP, WavLM, and RoBERTa to extract features from the visual, speech, and text modalities, respectively. To ensure consistency in feature dimensions across modalities, we apply a linear transformation layer to the speech features. These extracted features are then passed through a Feature Extractor, which consists of a single-layer Transformer encoder designed to refine contextual representations and enhance the quality of the learned embeddings. The resulting preliminary modality representations are formulated as:  
\vspace{-1mm}
\begin{equation}
    H_V \in \mathbb{R}^{k \times d}, \quad H_S \in \mathbb{R}^{m \times d}, \quad H_T \in \mathbb{R}^{n \times d},
\end{equation}
where \( k, m, n \) denote the sequence lengths of the video, speech, and text modalities, respectively, and \( d \) represents the unified feature dimension.

\subsection{Multimodal Interaction Module}
Our multimodal interaction module consists of MSR and MIR modules. We describe these two modules in detail below.

\subsubsection{Modality-specific Representations (MSR) Module}
The MSR module is designed to capture modality-specific emotional features while dynamically regulating intermodal interactions. To achieve this, we introduce the gated interactive attention (GIA) mechanism, which adaptively controls the information flow between different modalities.

\noindent \textbf{GIA Block.} The GIA mechanism enables adaptive information fusion between modalities by learning modality-specific gating factors. Given two modalities \( A \) and \( B \), the cross-attention mechanism computes the attention-weighted representation of \( A \) based on \( B \):
\vspace{-2mm}
\begin{equation}
    H_{A \rightarrow B} = {\rm Softmax} \left( \frac{Q_A K_B^\top}{\sqrt{d}} \right) V_B \in \mathbb{R}^{t_A \times d},
\end{equation}
where \( Q_A = W_Q H_A \in \mathbb{R}^{t_A \times d} \), \( K_B = W_K H_B \in \mathbb{R}^{t_B \times d} \), and \( V_B = W_V H_B \in \mathbb{R}^{t_B \times d} \) are the query, key, and value matrices for modalities \( A \) and \( B \), with \( t_A \) and \( t_B \) as their sequence lengths, respectively. Here, \( W_Q \in \mathbb{R}^{d \times d} \), \( W_K \in \mathbb{R}^{d \times d} \), and \( W_V \in \mathbb{R}^{d \times d} \) are learnable weight matrices for the query, key, and value transformations.

To control the information flow, we introduce a modality-specific gating function, which learns to regulate the influence of modality \( B \) on modality \( A \):


\vspace{-3mm}
\begin{equation}
    G = \sigma(W_g ({\rm Norm}({\rm AvgPooling}(H_{A \rightarrow B})) + b_g)) \in \mathbb{R}^{t_A \times d},
\end{equation}
where \( W_g \in \mathbb{R}^{d \times d} \) and \( b_g \in \mathbb{R}^{d} \) are learnable parameters, “\( \sigma(\cdot) \)” is the sigmoid activation function.
The final MSR of modality A is computed as:

\vspace{-3mm}
\begin{equation}
    H_{A}^{\rm (GIA)_\textit{B}} = G \odot H_{A \rightarrow B} + (1 - G) \odot H_A  \in \mathbb{R}^{t_A \times d},
\end{equation}
where “\( \odot \)” represents element-wise multiplication. Similarly, we can obtain the gated representation $H_{B}^{\rm (GIA)_\textit{A}} \in \mathbb{R}^{t_B \times d}$ for modality \( B \).
The MSR module is composed of three GIA blocks, each designed to capture pairwise interactions between modalities. Specifically, the interactions are formulated as:
\begin{equation}
\begin{aligned}
    & \{ H_V^{\rm (GIA)_\textit{S}}, H_S^{\rm (GIA)_\textit{V}} \} = \text{GIA}(H_V, H_S), \\
    & \{ H_S^{\rm (GIA)_\textit{T}}, H_T^{\rm (GIA)_\textit{S}} \} = \text{GIA}(H_S, H_T), \\
    & \{ H_T^{\rm (GIA)_\textit{V}}, H_V^{\rm (GIA)_\textit{T}} \} = \text{GIA}(H_T, H_V),
\end{aligned}
\end{equation}
where $H_V^{\rm (GIA)_\textit{S}}, H_V^{\rm (GIA)_\textit{T}} \in \mathbb{R}^{k \times d}$ , $H_S^{\rm (GIA)_\textit{V}}, H_S^{\rm (GIA)_\textit{T}} \in \mathbb{R}^{m \times d}$, $H_T^{\rm (GIA)_\textit{S}}, H_T^{\rm (GIA)_\textit{V}} \in \mathbb{R}^{n \times d}$.
For each modality, we obtain its final MSR by summing the outputs from the corresponding GIA blocks:
\begin{equation}
\begin{aligned}
    & H_V^{\rm (MSR)} = H_V^{\rm (GIA)_\textit{S}} + H_V^{\rm (GIA)_\textit{T}} \in \mathbb{R}^{k \times d}, \\
    & H_S^{\rm (MSR)} = H_S^{\rm (GIA)_\textit{V}} + H_S^{\rm (GIA)_\textit{T}} \in \mathbb{R}^{m \times d}, \\
    & H_T^{\rm (MSR)} = H_T^{\rm (GIA)_\textit{V}} + H_T^{\rm (GIA)_\textit{S}}\in \mathbb{R}^{n \times d}.
\end{aligned}
\end{equation}
The overall MSR is obtained by concatenating the MSR of each modality:

\vspace{-2mm}
\begin{equation}
    H^{\rm (MSR)} = [H_V^{\rm (MSR)}, H_S^{\rm (MSR)}, H_T^{\rm (MSR)}] \in \mathbb{R}^{(k+m+n) \times d},
\end{equation}
which ensures that each modality retains its distinct emotional features while benefiting from cross-modal interactions.

\subsubsection{Modality-invariant Representations (MIR) Module}
The MIR module aims to extract shared emotional features across different modalities, ensuring that the learned representations capture modality-agnostic emotional cues. Unlike the MSR module, which emphasizes modality-dependent features, the MIR module focuses on learning a common latent space where multimodal features are aligned. The MIR module consists of three modality-invariant generator (MIG) blocks, each is responsible for refining the representation of a specific modality by incorporating cross-modal interactions.

\noindent \textbf{MIG block.} To establish a shared interaction space, we concatenate the preliminary embeddings from all three modalities to form the query $Q$:

\vspace{-2mm}
\begin{equation}
H_{VST} = [H_V, H_S, H_T] \in \mathbb{R}^{(k+m+n) \times d}.
\end{equation}
For each modality \( M \in \{V, S, T\} \), we use the corresponding \( H_M^{\rm (MSR)} \) as the key $K$ and value $V$. Cross-modal attention is then applied to compute inter-modal interactions:  

\vspace{-2mm}
\begin{equation}
    H_M^{\rm(share)} = {\rm Softmax} \left( \frac{Q K^\top}{\sqrt{d}} \right) V \in \mathbb{R}^{(k+m+n) \times d}.
\end{equation}
To enhance the feature's modality invariance, a parallel convolutional network is employed to learn a mask that filters out modality-specific information:
\begin{equation}
\begin{aligned}
H_M^{\rm(b)} = H_M^{\rm(share)}& \otimes \sigma ({{\rm Conv_{1d}}([H_M^{\rm(MSR)} , H_{VST}])}) \\
&\in \mathbb{R}^{(k+m+n) \times d}, \quad M \in \{V, S,T\},
\end{aligned}
\end{equation}
where “$\otimes$” indicates element-wise multiplication and “${\rm Conv_{1d}}$” denotes $1 {\rm \times} 1$ convolution followed by PReLU activation \cite{he2015delving}.
To further refine the modality-invariant representation, we introduce an additional 1D convolutional layer with a stride of 2 that enhances local feature extraction and interaction, helping to capture fine-grained dependencies across modalities. Additionally, we incorporate a residual connection to preserve the original multimodal information:
\begin{equation}
\begin{aligned}
H_M^{\rm(MIG)} =& {\rm Norm} \left( H_{VST} + {\rm Conv_{1d}}  \left( H_M^{\rm(b)} \ \right) \right)
\\
&\in \mathbb{R}^{(k+m+n) \times d}, \quad M \in \{V, S,T\},
\end{aligned}
\end{equation}
where “{\rm Norm}” represents layer normalization \cite{ba2016layer}. The overall MIR is obtained by concatenating the MIR of each modality:
\vspace{-1mm}
\begin{equation}
    H^{\rm (MIR)} = [H_V^{\rm (MIG)}, H_S^{\rm (MIG)}, H_T^{\rm (MIG)}] \in \mathbb{R}^{3(k+m+n) \times d},
\end{equation}

Finally, the MSR and MIR are concatenated together to obtain the final multimodal interactive  representation $H_{VST}^{\rm(fus)}$:
\begin{equation}
H_{VST}^{\rm(fus)} = H^{\rm (MSR)} + H^{\rm (MIR)} \in \mathbb{R}^{4(k+m+n) \times d},
\end{equation}
\noindent \textbf{Modality-Invariant Learning Constraints (MIC).}  
To enforce constraints on the consistency of modality-invariant representations across different modalities, we utilize the symmetric KL divergence (SKL) to obtain a more stable and bidirectional similarity measure. The SKL divergence between two modality-invariant representations \( H_M^{\rm (MIR)} \) and \( H_N^{\rm (MIR)} \) is computed as follows:  
\vspace{-6mm}

\begin{equation}
\begin{aligned}
D_{\rm SKL}(H_M^{\rm (MIR)}, H_N^{\rm (MIR)}) &= \frac{1}{2} \left( D_{\rm KL}(H_M^{\rm (MIR)} \parallel H_N^{\rm (MIR)}) \right. \\ 
&\quad + \left. D_{\rm KL}(H_N^{\rm (MIR)} \parallel H_M^{\rm (MIR)}) \right),
\end{aligned}
\end{equation}
where \( M, N \in \{V, S, T\} \) represent the three modalities, and \( D_{\rm KL}(P \parallel Q) \) is defined as:

\begin{equation}
    D_{\rm KL}(P \parallel Q) = \sum_i P(i) \log \frac{P(i)}{Q(i)}.
\end{equation}
To enforce cross-modal alignment, we compute the SKL divergence for all modality pairs:  
\begin{equation}
\begin{aligned}
&\mathcal{L}_{\rm MIR} = D_{\rm SKL}(H_V^{\rm (MIR)}, H_S^{\rm (MIR)}) + \\
&D_{\rm SKL}(H_S^{\rm (MIR)}, H_T^{\rm (MIR)}) + D_{\rm SKL}(H_T^{\rm (MIR)}, H_V^{\rm (MIR)}).
\end{aligned}
\end{equation}

This loss function encourages the MIR distributions of different modalities to be similar, thereby improving the consistency of cross-modal representations. By minimizing \( \mathcal{L}_{\rm MIR} \), we ensure that the learned MIR representations reside in a shared latent space, facilitating more effective multimodal fusion and enhancing the overall performance of emotion recognition.
\vspace{-3mm}

\subsection{Emotion Classification Module}
The emotion classification is performed by applying the temporal average pooling layer on the concatenated MSR and MIR features $H_{VST}^{\rm (fus)}$ followed by a linear layer and a Softmax activation function.
%
\begin{equation}
\label{eq13}
  P(\hat{l}|H_{VST}^{\rm(fus)}) = {\rm Softmax}((W_c({\rm AvgPooling}(H_{VST}^{\rm(fus)}))+ b_c)),
\end{equation}
where \( W_c \) and \( b_c \) are learnable parameters and $\hat{l}$ is the predicted emotion classification.  The corresponding loss function can be defined as
\begin{equation}
\mathcal{L}_{\rm ER} = - \sum{\rm log}(P(\hat{l}|H_{VST}^{\rm(fus)})).
\end{equation}
\noindent \textbf{Joint Training.}
During the training stage, the two loss functions are linearly combined as the overall training objective:
\begin{equation}
\mathcal{L} = \mathcal{L}_{\rm ER} + \gamma  \mathcal{L}_{\rm MIR},
\end{equation}
where $\gamma$ is the hyperparameter for balancing the weight of two loss functions.

\vspace{-3mm}
\section{Experiments and Results}
\subsection{Experimental Conditions}
\noindent \textbf{Experiment Settings.} Our method was implemented with Python 3.10.0 and Pytorch 1.11.0 and was trained on a system with an Intel Xeon Gold 6248 CPU, 32GB RAM, and an NVIDIA Tesla V100 GPU. The visual, speech, and text encoders were initialized using CLIP\footnote{\href{https://huggingface.co/openai/clip-vit-large-patch14}{https://huggingface.co/openai/clip-vit-large-patch14}}, WavLM\footnote{\href{https://huggingface.co/microsoft/wavlm-large}{https://huggingface.co/microsoft/wavlm-large}}, and RoBERTa\footnote{\href{https://huggingface.co/FacebookAI/roberta-base}{https://huggingface.co/FacebookAI/roberta-base}}, producing feature representations of 768, 1,024, and 768 dimensions, respectively. A linear layer mapped speech features to 768 dimensions. During training, CLIP and RoBERTa were kept frozen, while WavLM was fine-tuned. The speech modality input consisted of 6 seconds of audio at a 16 kHz sampling rate. Given the presence of two speakers in the video, speaker separation was required. Ultimately, the visual modality input comprised 180 individual images. The feature extractor comprised a single-layer transformer with a hidden size of 768. We optimized using Adam \cite{kingma2014adam} with a batch size of 32, a fixed learning rate of $1e^{-5}$, and $\gamma$ set to 0.1 in the loss function.

\noindent \textbf{Dataset.} To demonstrate the effectiveness of our proposed method, we conducted experiments on the IEMOCAP dataset \cite{busso2008iemocap}. This dataset comprised roughly 12 hours of audio, video, transcriptions, and motion-capture information from ten speakers in five scripted sessions. Following prior work, we employed 5,531 utterances from four emotion categories: “neutral”, “angry”, “happy”, and “sad”, with “excited” merged into the “happy” category. We conducted experiments with 5-fold leave-one-session-out cross-validation.

\vspace{-3mm}
\subsection{Results and Analysis}

\noindent \textbf{Comparisons of the SOTA Methods.} Table \ref{tab:results_IEMOCAP} compares the performance of the GIA-MIC method with recent multimodal SER approaches on IEMOCAP. Our proposed GIA-MIC achieves a weighted accuracy (WA) of 80.7\% and an unweighted accuracy (UA) of 81.1\% on IEMOCAP, outperforming other SOTA methods. We used Whisper \cite{radford2022robust} to generate ASR transcripts, achieving a word error rate (WER) of 20.48\% on the IEMOCAP dataset. Even with ASR transcripts, GIA-MIC outperforms other methods, further demonstrating its effectiveness.

\begin{table}[ht]
\centering
\caption{Performance Comparison of SOTA Models on IEMOCAP (\%). ``S", ``T" and ``V" represent speech, text and visual modalities, respectively. ``ASR'' and ``GT" denote ASR and ground truth transcripts, respectively. The best and second-best results are marked in \textbf{bold} and \underline{underlined}, respectively.}
\vspace{-3mm}
\label{tab:results_IEMOCAP}
\resizebox{\linewidth}{!}{ 
\begin{tabular}{l|c|c|cc}
\toprule
\textbf{Method} & \textbf{Year} & \textbf{Modality} & \textbf{WA} & \textbf{UA} \\
\midrule
\midrule
RMSER-AEA \cite{lin2023robust} & 2023 & S+T(ASR) & 76.4 & 76.9 \\
MGAT \cite{fan2023mgat} & 2023 & S+T(GT) & 78.5 & 79.3 \\
IMISA \cite{wang2024inter} & 2024 & S+T(GT) & 77.4 & 77.9 \\
MFLA \cite{shi2024study} & 2024 & S+T(ASR) & - & 77.7 \\
MF-AED-AEC \cite{he2024mf} & 2024 & S+T(ASR) & 78.1 & 79.3 \\
MAF-DCT \cite{wu2024multi} & 2024 & S+T(GT) & 78.5 & 79.3 \\
FDRL \cite{sun2024fine} & 2024 & S+T(GT) & 78.3 & 79.4 \\

CAT-BC \cite{fan2025coordination} & 2025 & S+T(GT) & \underline{79.5} & \underline{80.3} \\
MCWSA-CMHA \cite{zheng2022multi} & 2022 & V+S & 78.9 & - \\
GCNet \cite{lian2023gcnet}  & 2023 & V+S & 78.4 & - \\
Foal-Net \cite{li2024enhancing}  & 2024 & V+S & 79.5 & 80.1 \\
S2MER-CMDM \cite{liang2020semi} & 2020 & V+S+T(GT) & 75.6 & 74.5 \\
AMED \cite{heredia2022adaptive} & 2022 & V+S+T(GT) & - & 77.6 \\
FM-MER \cite{pena2023framework} & 2023 & V+S+T(GT) & - & 78.9 \\
\midrule
\rowcolor[HTML]{EFEFEF} 
\textbf{GIA-MIC (Ours)} & \textbf{2025} & \textbf{V+S+T(ASR)} & \textbf{79.6} & \textbf{80.3} \\
\rowcolor[HTML]{EFEFEF} 
\textbf{GIA-MIC (Ours)} & \textbf{2025} & \textbf{V+S+T(GT)} & \textbf{80.7} & \textbf{81.3} \\
\bottomrule
\end{tabular}
} 
\end{table}

\noindent \textbf{Ablation Studies.} 
We conduct ablation studies to evaluate the contributions of the MSR, MIR, and MIC modules in our GIA-MIC framework. Results in Table \ref{tab:as_IEMOCAP} show that removing any component causes a performance drop, emphasizing their role in enhancing MER.
Among the three, MSR has the greatest impact, with its removal resulting in a 1.1\% WA and 0.9\% UA decrease, highlighting the importance of MSR information for emotion recognition. The MIR module also proves vital, with a 0.6\% WA drop when removed, suggesting its role in cross-modal alignment. Although MIC has the smallest effect, removing it leads to a 0.5\% WA decrease, indicating its contribution to refining learned representations through consistency across modalities.
Overall, the full GIA-MIC model outperforms the “Baseline” approach, confirming that combining MSR, MIR, and MIC is crucial for effective multimodal fusion.

\begin{table}[htbp]
    \centering
    \caption{The results of ablation experiments on IEMOCAP (\%).}
    \vspace{-3mm}
    \label{tab:as_IEMOCAP}
    \begin{tabular}{c|c|cc}
        \toprule
        \textbf{Modality} & \textbf{Method} & \textbf{WA} & \textbf{UA} \\
        \midrule
        \midrule
        V & CLIP & 46.6 & 47.9 \\
        S & WavLM & 71.7 & 72.4 \\
        T(GT) & RoBERTa & 68.4 & 69.4 \\
        \midrule
        \multirow{1}{*}{V+S} & Baseline & 72.3 & 73.1 \\
        \multirow{1}{*}{S+T(GT)} & Baseline & 76.4 & 77.2 \\
        \multirow{1}{*}{V+T(GT)} & Baseline & 69.0 & 69.9 \\
        \multirow{1}{*}{V+S} & GIA-MIC (ours) & 79.4 & 80.1 \\
        \multirow{1}{*}{S+T(GT)} & GIA-MIC (ours) & 79.0 & 80.4 \\
        \multirow{1}{*}{V+T(GT)} & GIA-MIC (ours) & 71.3 & 72.0 \\
        \midrule
        \rowcolor[HTML]{EFEFEF} 
        \multirow{1}{*}{\textbf{V+S+T(GT)}} & \textbf{GIA-MIC (ours)} & \textbf{80.7} &\textbf{ 81.3} \\
        \hspace{1em} w/o MSR & GIA-MIC (ours) & 79.6 & 80.4 \\
        \hspace{1em} w/o MIR & GIA-MIC (ours) & 80.1 & 80.7 \\
        \hspace{1em} w/o MIC & GIA-MIC (ours) & 80.2 & 80.9 \\
        \bottomrule
    \end{tabular}
\end{table}

\begin{figure}[t]
  \centering
  \includegraphics[width=1\columnwidth]{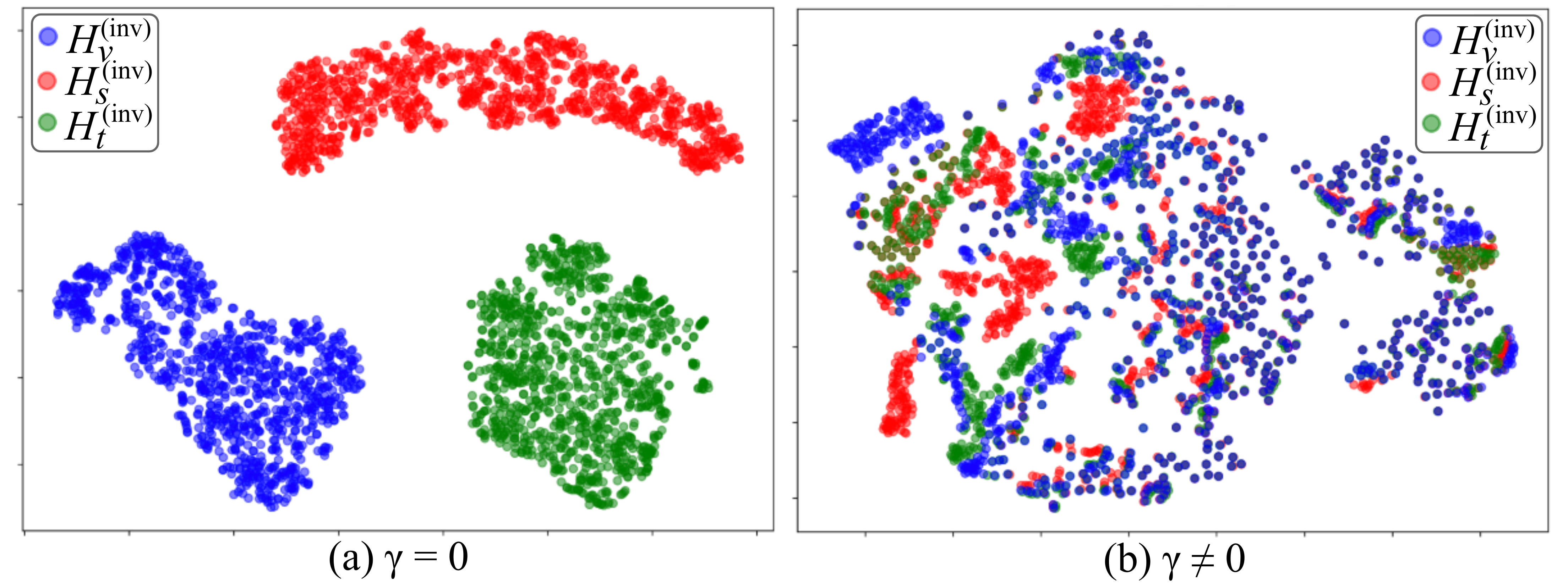}
  \vspace{-6mm}
  \caption{t-SNE visualizations of the distribution of modality-invariant representations before and after introducing the modality-invariant  learning constraints.}
  \vspace{-6mm}
  \label{fig:tsne}
\end{figure}


\noindent \textbf{Visualization Analysis.} We employ t-SNE \cite{van2008visualizing} to visualize the multimodal modality-invariant representations, as shown in Fig. \ref{fig:tsne}. When $\gamma = 0$, meaning no modality-invariant constraints are applied, the representations of different modalities remain distinct. In contrast, with temporal modality-invariant constraints ($\gamma \neq 0$), the three modalities exhibit greater overlap, indicating increased shared information. This suggests that the constraints effectively enhance modality alignment, leading to more similar representations after training.

\section{Conclusion}
In this paper, we propose GIA-MIC, a novel multimodal emotion recognition framework that effectively integrates modality-invariant and modality-specific representations while enforcing cross-modal consistency through modality-invariant constraints. Our approach addresses the challenges of modality heterogeneity and misalignment by learning both shared and distinct representations, enabling a more robust fusion of visual, speech, and text modalities.
Extensive experiments on the IEMOCAP dataset demonstrate that GIA-MIC significantly outperforms baseline methods, achieving SOTA performance.

\section{Acknowledgements}
This work was partly supported by JST AIP Acceleration Research JPMJCR25U5 and JSPS KAKENHI Grant Number 21H05054, Japan.


\bibliographystyle{IEEEtran}
\bibliography{mybib}

\end{document}